\documentclass[letterpaper, 10pt, conference]{ieeeconf}

\IEEEoverridecommandlockouts

\usepackage{graphicx} 
\usepackage{amsmath} 
\usepackage{amssymb}  
\usepackage{commath}

\usepackage{pifont}
\newcommand{\cmark}{\ding{51}}%
\newcommand{\xmark}{\ding{55}}%
\usepackage{textcomp}

\usepackage{lineno}
\usepackage{tikzpagenodes}
\usepackage{background}
\usepackage{balance}
\usepackage{multirow}
\usepackage{balance}

\usepackage{color}

\definecolor{todocolor}{rgb}{0.7,0.0,0.1}

\definecolor{todiscusscolor}{rgb}{0.1,0.0,0.7}

\definecolor{timcolor}{rgb}{0.1,0.7,0.0}

\backgroundsetup{color=white}

\title{\LARGE \bf
Addressing the Sim2Real Gap in Robotic 3D Object Classification
}

\author{Jean-Baptiste Weibel, Timothy Patten and Markus Vincze
\thanks{The research leading to these results has received funding from the Austrian Science Foundation (FWF) under grant agreement No. I3968-N30 HEAP and No. I3969-N30 InDex}
\thanks{All authors are with the Vision for Robotics Laboratory, Automation and Control Institute, TU Wien, 1040 Vienna, Austria.
{\tt\small\{weibel, patten, vincze\}@acin.tuwien.ac.at}}%
}

\begin{document}

\maketitle

\begin{abstract}
Object classification with 3D data is an essential component of any scene understanding method. It has gained significant interest in a variety of communities, most notably in robotics and computer graphics. While the advent of deep learning has progressed the field of 3D object classification, most work using this data type are solely evaluated on CAD model datasets. Consequently, current work does not address the discrepancies existing between real and artificial data. In this work, we examine this gap in a robotic context by specifically addressing the problem of classification when transferring from artificial CAD models to real reconstructed objects. This is performed by training on ModelNet (CAD models) and evaluating on ScanNet (reconstructed objects). We show that standard methods do not perform well in this task. We thus introduce a method that carefully samples object parts that are reproducible under various transformations and hence robust. Using graph convolution to classify the composed graph of parts, our method significantly improves upon the baseline. \\\noindent-- Code will be made publicly available on acceptance. --

\end{abstract}

\section{Introduction}

Whether to recommend the most suitable CAD model to a designer or to enable a service robot to decide where to place objects when tidying a room, 3D object classification is an essential task. Research in this area has greatly benefited from the wide availability of 3D CAD models as well as the accessibility of depth sensors, as this has established a large amount of data to apply geometric reasoning.

Deep learning has profoundly transformed computer vision in recent years, and in particular, object classification has seen spectacular improvements. There has been steady interest for applying these methods for 3D data but introducing geometric reasoning in deep learning is not without its pitfalls. Typical deep learning approaches cannot handle rotated objects and real-world objects might be observed in arbitrary poses. Some methods use the statistical distribution of the data to transform the unknown object to a canonical pose for the deep network~\cite{jaderberg_spatial_2015, qi_pointnet:_2017}.
However, inaccessible viewpoints, partial occlusions, supporting surfaces, and over- or under-segmentation observed in real-world data all contribute to modifying the statistical distribution of data that these methods expect, thus hindering their performance. Most deep networks also expect a fixed size input. This is achieved by rescaling, however, applying this to an occluded object can lead to a significant difference in the final fixed size representation. Consider how rescaling and centering a model airplane to the unit sphere and the same model with one wing missing would produce vastly different coordinates. The effect becomes prevalent when transferring from CAD models to real-world objects, as scale information is not available during training since most CAD models are scaleless.

\begin{figure}[t]
   \centering
   \includegraphics[scale=0.35]{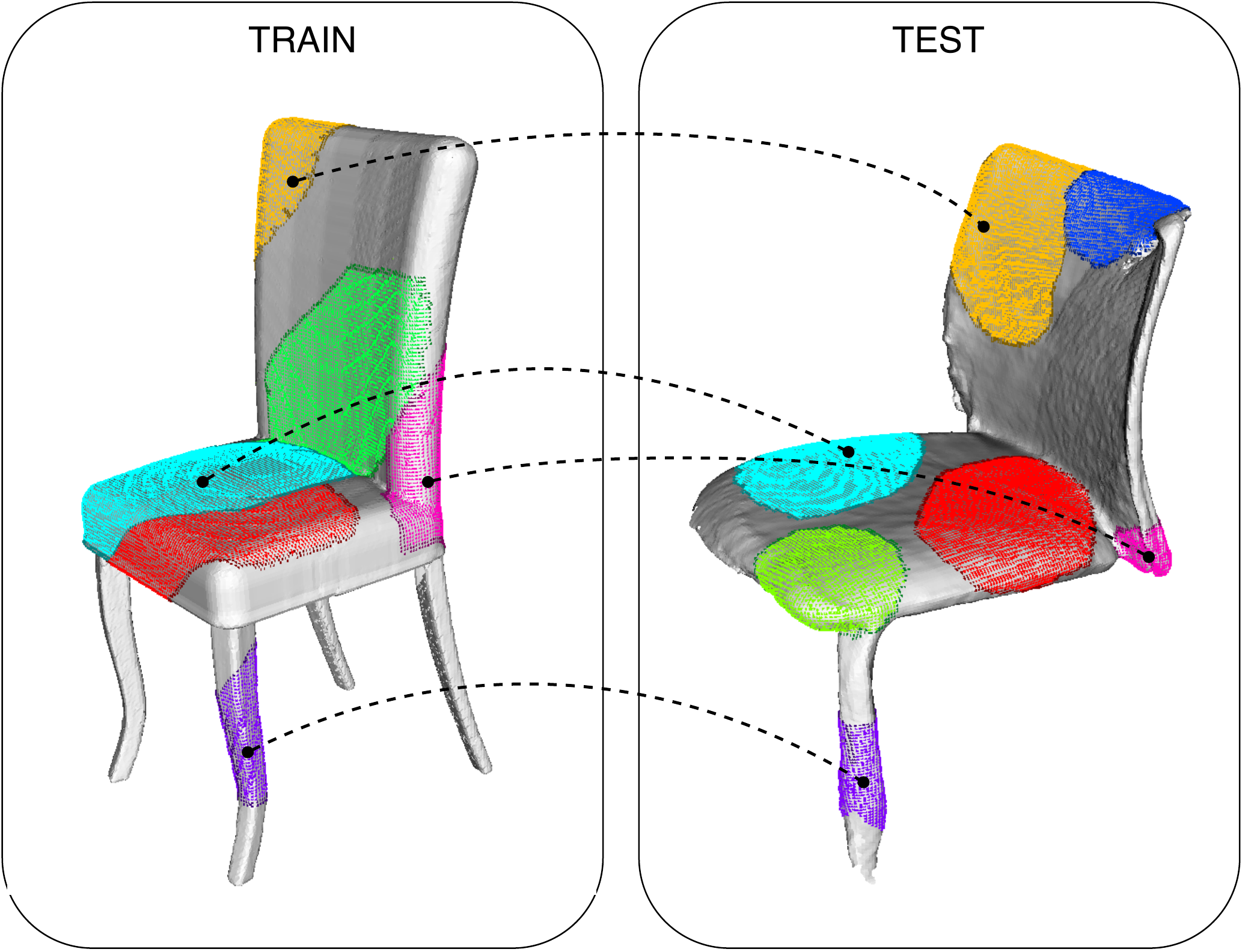}
   \caption{Creating reproducible object parts with similar representations on all sources of data enables better transfer from artificial to real objects.}
   \label{fig:noise}
\end{figure}

In this work, we develop a method for 3D object classification based on object parts that are reproducible under orientation or scale changes and can be defined for any level of occlusion, as shown in Figure~\ref{fig:noise}. It should be noted that we define parts as a \emph{continuous subset} of the original object without any specific semantic meaning. Indeed, semantic parts such as a cup handle or a chair leg are also likely to be occluded and impossible to recover from the original objects, whereas our non-semantic parts can always be defined. Once parts are extracted, a rotation-invariant representation is computed through the use of a reproducible local reference frame. Finally, a graph-convolution based architecture is used to classify the graph of parts.

In summary, our contributions are the introduction of:
\begin{enumerate}
    \item a carefully designed angle-based sampling procedure that creates object parts reproducible under various rotation, scale and occlusion and,
    \item a general graph-based learning architecture for classification that preserves the relevant properties of 3D object parts.
\end{enumerate}
These aspects allow us to achieve high performance when transferring from artificial to real-world data. In particular, our approach transfers from the ModelNet dataset~\cite{MODELNET} to objects segmented from the reconstructed scenes in the ScanNet dataset~\cite{dai2017scannet} better than previous methods. Our approach also significantly outperforms state-of-the-art methods when training and testing on ScanNet, which further demonstrates the value of careful part design and the inclusion of geometric priors.

The remainder of the paper is organized as follows. Section~\ref{sec:related_work} reviews the relevant literature. Section~\ref{sec:method} describes our approach for creating a graph of parts and subsequent learning for object classification. Section~\ref{sec:experiments} presents results of our method in comparison to existing methods, both on artificial data, real data, and when transferring from the former to the latter. Finally, Section~\ref{sec:conclusion} concludes and discusses future work.

\section{Related Work}\label{sec:related_work}

This section will first discuss approaches for geometric deep learning, and then approaches specifically designed for noisy data.

\subsection{Geometric Deep Learning}

Due to the way cameras perceive the world, 3D object classification is mostly concerned with surfaces (i.e. a 2D manifold in 3D space) rather than volumes. These surfaces are best represented with meshes, which is a specific type of graph.
The branch of deep learning concerned with such data has recently received much attention from fields such as computer graphics as it focuses on triangle meshes. Graph convolutional models were originally designed for citation graphs and other graph-structured high-level data~\cite{kipf2017semi}, but more complex models have since developed for object meshes~\cite{verma_feastnet:_2018, SurfNet_2018_CVPR}. These models are typically developed for and tested on CAD models, which allows certain design choices such as using vertex coordinates. Unfortunately coordinates change significantly with rotation or rescaling and therefore these methods are not suitable for the unpredictability of real-world data.

High performance is achieved by taking advantage of the progress made in 2D classification by using a collection of 2D views of the object. They either pool over views~\cite{su15mvcnn, 10.1007/978-3-030-11015-4_49}, apply more advanced schemes such as intelligently clustering before pooling~\cite{wang2017dominant} or jointly learning the corresponding object pose~\cite{kanezaki2018_rotationnet}. Beyond the power of such representations, they also take advantage of networks pre-trained on large scale 2D datasets. Once again, however, they have been designed and tested only on CAD models. Experiments suggest they heavily depend on the outline of the object in the 2D view, which is significantly affected by occlusions~\cite{10.1007/978-3-030-11015-4_49}.

Another option for 3D data is to apply 3D convolution on voxel grids~\cite{MODELNET, maturana2015voxnet}. 3D fixed grids are, by design, very sensitive to differences in object orientation and occlusion. It is possible to train with objects in a variety of poses~\cite{SZB17a, zeng20163dmatch}, but this amounts to learning as many representations as orientations. Resultingly, the methods require a larger number of parameters to attain a given accuracy. Working with an additional dimension compared to images also leads to the parameter count increasing much faster. The trade-off between the coarseness of the grid and model complexity inherent to this representation led to the exploration of more powerful fixed 3D grid representations such as the signed distance function~\cite{Park_2019_CVPR}. The limitation can also be counterbalanced by using multiple resolutions~\cite{Riegler2017OctNet}. KD-tree-based models~\cite{Klokov_2017_ICCV} push the multi-resolution idea further. Learning from this data structure achieves high accuracy but the KD-tree itself is sensitive to sensor noise and slight rotations, which makes it unsuitable for real sensor data.

Depth sensors sample perceived surfaces and provide point cloud data that is used for direct learning. PointNet~\cite{qi_pointnet:_2017} exploits this data type by learning from one point at a time before using a global max pooling layer to optimize globally over all the positions in the unit sphere, independently of the order of the points in the set. Research is still very active in this area with methods creating local kernels~\cite{thomas2019KPConv, Tat2018} or exploring novel classifiers~\cite{DBLP:journals/corr/abs-1811-02191}.


\subsection{Robust 3D Classification}

The methods mentioned until now are evaluated using the coordinates of the models in the unit sphere. Objects extracted from a reconstructed scene, however, come in any orientation, which is often detrimental to the performance of methods that ignore the difference. Common approaches to this challenge are to train over various orientations as in~\cite{maturana2015voxnet} or to use a spatial transformer layer~\cite{jaderberg_spatial_2015} to learn an alignment of the objects as in~\cite{qi_pointnet:_2017}.

Another direction is to learn from rotation-invariant features as is explored in \cite{Chen_2019_CVPR, 8794432, eppf18}. While a new representation is introduced in~\cite{Chen_2019_CVPR}, most work take inspiration from classical feature descriptors and explore the potential combination with deep networks. For example, \cite{eppf18} takes inspiration from the Point Pair Features (PPF) descriptor~\cite{drost_model_2010} and use convolutional neural networks to learn a multi-dimensional histogram without losing the correlation between the features. The ESF descriptor~\cite{wohlkinger_ensemble_2011} is another handcrafted descriptor and is designed specifically for classification. It concatenates histograms over those sampled features, which is given to an SVM to classify. \cite{8794432} combines some of the features from ESF with PPF to learn local structures that are later combined with a graph convolutional network.

\section{Learning from Object Parts for Robust 3D Object Classification}\label{sec:method}

\begin{figure*}[t]
   \centering
   \includegraphics[scale=0.7]{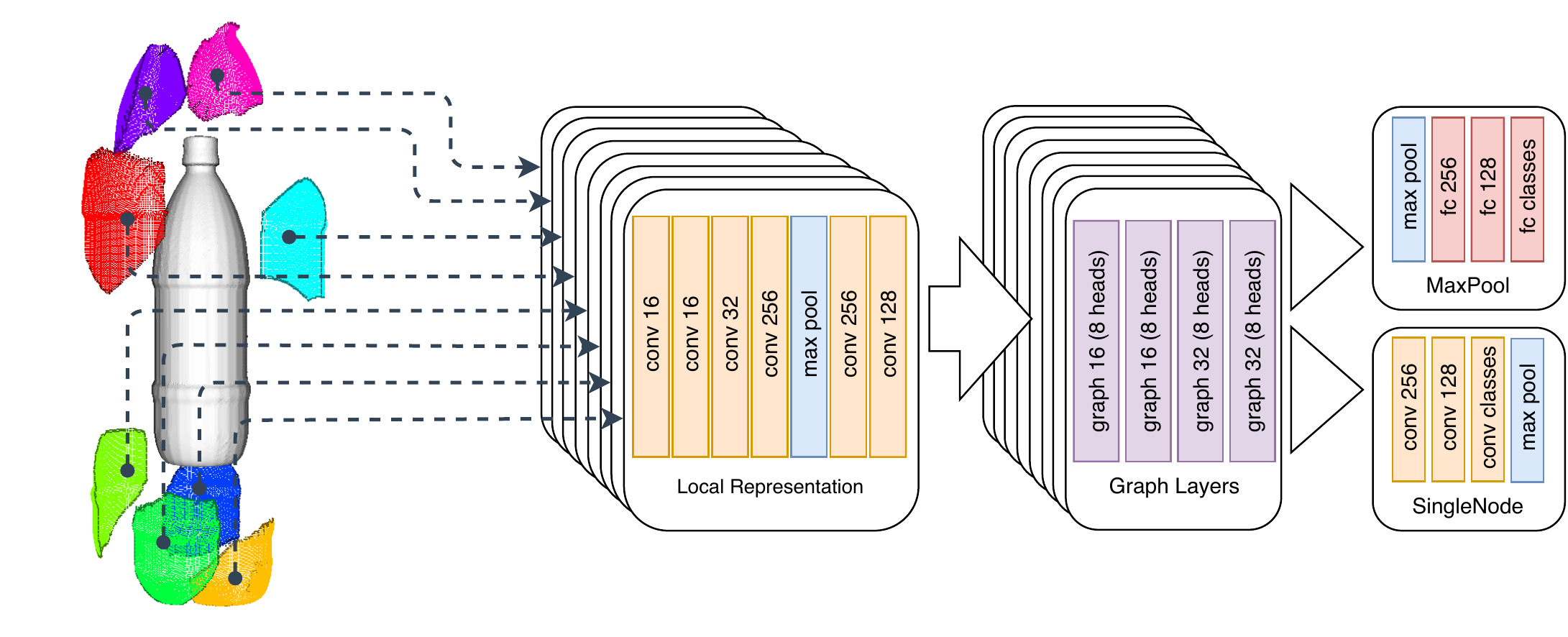}
   \caption{Architecture overview of our proposed method. The number of parts is reduced to eight for readability purposes and the connectivity is not displayed. Convolution layers all consider only a single element (kernel size one).}
   \label{fig:system}
\end{figure*}

Our method is developed for over- or under-segmented objects represented by manifold triangle meshes (a mesh is a manifold if each edge is connecting at most two triangles). The reason for using a mesh is that surfaces are preserved. Reconstruction methods generate that representation either directly~\cite{Schreiberhuber2019} or by applying a post-processing step such as the marching cube algorithm~\cite{InfiniTAM_ECCV_2016}. 
However, the method presented here could easily be adapted for dense point clouds by using nearest neighbor approaches to retrieve the neighborhood of each point.

This section describes the proposed method. We first explain the object part sampling process and the part representation. We then outline the learning approach for the graph of object parts and the design of our graph convolution architecture.

\subsection{Creating object parts}

\subsubsection{Object parts sampling}

To transfer from artificial object models to real reconstructed data, object parts should be repeatable under varying orientation, occlusions, scale and point density. Scale-invariance forbids the use of Euclidean distance for sampling parts. To avoid sampling a part that would span through an object, and thus being significantly more sensitive to occlusions, parts are grown by following the surface of the object. The average angle between a triangle and its neighbors on a surface is used when deciding whether a neighboring triangle should be added to the part. Since reconstruction algorithms account for sensor noise and artificial data does not suffer from any random noise, high quality normals can be computed for both type of data. The angle between two neighboring normals is independent of scale and orientation of the object, and in a perfectly noiseless case, even independent of the surface sampling density. Object parts are then extracted by performing a Breadth-First Search (BFS) on the graph defined by the object mesh, or in other words, incrementally adding a one-ring neighborhood around the sampled part center. Due to the strong unpredictability of occlusions, part centers are randomly sampled. Centers are sampled so long as they do not belong to a previously sampled part.  
The search is stopped when the accumulated angle over the object part reaches a set threshold. The accumulated angle is computed from the average angle of each triangle, which is simply the average of the angle with each triangle neighbor.

Reconstructed scenes do not provide perfectly smooth surfaces, therefore, we perform low-pass filtering on the normals defined by the triangles. Normals for all points are first computed by averaging the normals of each triangle they belong to. Then triangle normals are computed by averaging the normals of the three points. The resulting normals are smoother than the original mesh.

\subsubsection{Object part features}
The object part representation should maintain the properties of the sampling. In this work, we sample a fixed number of points from the object part to generate a fixed size representation from parts of varying sizes. Orientation-invariance is then achieved by defining a local reference frame (LRF). The center of the LRF is defined by the mean of a set of points and we propose two different orientations.

The first design option is to perform Principal Component Analysis (PCA) with the set of points and use the eigenvectors as the LRF. The first and last eigenvectors (when ordered by decreasing eigenvalues) are kept and the direction of the last eigenvector is flipped in order to follow the average direction of the surface normals of the set of points. This guarantees a different LRF for concave and convex sets. The last vector is the cross-product of the first two vectors. This LRF provides a total orientation invariance and is referred to as PCA-LRF.

The second option is to define the LRF based on the global vertical axis (Z-axis) and the component of the mean surface normal of the part that is orthogonal. This is no longer independent of the orientation of the object but only independent of the orientation of the object around the Z-axis. This local frame of reference is referred to as Z-LRF. Although it is only partially orientation-invariant, it offers a more informative representation. Since many objects have a small number of canonical poses (e.g. most bottles stand upright), it remains beneficial when tested on realistic data.

All point sets are rescaled to the unit sphere to make the representation independent of the scale. In most experiments in Section~\ref{sec:experiments}, we also add the average angle value as a feature to the point coordinates. When sampling the point, we use the average angle value of the triangle it belongs to. It slightly improves the accuracy without any extra computation overhead because it is computed during sampling.

Finally, the graph is constructed by connecting parts that overlap. In other words, parts are connected if at least one triangle in each of the parts was sampled in the original object.

\subsection{Model Architecture}

\subsubsection{General architecture}
The architecture, as shown in Figure~\ref{fig:system}, follows the PointNet model where each point extracted from a part is independently fed to the same convolutional neural network. In difference to the architecture of PointNet, the spatial transformer network is unnecessary as the points' coordinates are already defined in a LRF. Instead, four 1D convolutional layers are used with a kernel size of one. A progressively increasing number of filters are then applied before a max-pooling layer to pool over each object part. Each of the layers includes a batch normalization step~\cite{ioffe_batch_2015}. A weighted version of the maximum value (over the whole set) of a given filter is subtracted from each output as described in~\cite{iclr_2017_dl_sets_pc}. ReLU is used as an activation function.

The proposed model includes graph convolution layers inspired by the GCN model introduced in~\cite{kipf2017semi}. This is a simplification of larger graph convolutional models because only first-degree neighbors are considered. As a result, the model cannot differentiate neighbors from each other. To address this, we introduce an attention model in our graph convolutions.

\subsubsection{Attention model}
The GCN model is made more powerful by introducing an attention mechanism. Instead of considering all neighbors as equal, the neighbors are weighted according to a criterion that is specific to the attention model chosen. Examples of attention models have been developed in~\cite{verma_feastnet:_2018} and~\cite{velickovic2018graph}. To further improve the representational power of a network, multiple attention heads are used for the same layer, and the attention heads output are concatenated at each layer.

Introducing attention to the GCN model amounts to learning a valid coefficient to replace the normalization factor. We follow the model defined in \cite{velickovic2018graph} where the graph convolution layer becomes
\begin{equation}
h_{v_i}^{l+1} = \sigma \left ( \sum_{j \in \mathcal{N}_i} \gamma_{ij} h_{v_j}^l W^l \right ),
\end{equation}
where $h_{v_i}$ is the feature vector of the $i$-th vertex, $\sigma$ is the activation function, and $W$ is the parameter vector of the layer $l$. The coefficient $\gamma_{ij}$ is defined as
\begin{equation}
\begin{split}
\gamma_{ij} & = \text{softmax}(e_{ij}), \\
           & = \frac{\text{exp}(e_{ij})}{\sum_{k \in \mathcal{N}_i} \text{exp}(e_{ik})}, \\
\end{split}
\end{equation}
where $e_{ij} = \text{LeakyReLU}(a^T . [h_{v_i} W || h_{v_j} W]))$, $(\cdot)^T$ denotes the transpose operation, $\cdot||\cdot$ denotes the concatenation operation and $a$ is the vector of learned parameters for the attention. In order to respect the part connectivity, we add a bias matrix to the $e_{ij}$ term before applying the softmax in which disconnected nodes have a value of $-10^{9}$ and connected nodes have a value of $0$.

\subsubsection{Summarizing over object parts}
In this work, we are interested in predicting the object-level class. The object part representation described so far affords a number of different options for this task. The most straightforward option is to simply perform a max-pooling operation on the feature vectors of each part and then classifying the object (referred to as MaxPool). However, the classifier will be trained expecting all nodes and is therefore less likely to transfer well to real reconstructed data that have missing nodes. A second option is to predict one class per node and average all predictions into an object-level prediction (referred to as SingleNode). Both options are evaluated in Section~\ref{sec:experiments}. The single node prediction trained on artificial data still provides a representation that assumes perfect connectivity. We therefore propose one last option in which a proportion of nodes are randomly disconnected (except for self-connections).

\section{Experiments}\label{sec:experiments}
This section presents the experimental results. The first set of experiments compares our proposed approach to state-of-the-art methods for object classification on artificial data using the ModelNet dataset~\cite{MODELNET}. The second set of experiments evaluates the transfer abilities from artificial data (ModelNet) to real-world data (objects extracted from the ScanNet dataset~\cite{dai2017scannet}) in comparison to the baseline PointNet~\cite{qi_pointnet:_2017}. We also evaluate in more depth the impact of the object part size and the connectivity of the object parts graph. Lastly, our method is evaluated against the PointNet architecture when training and testing on real-world data with objects extracted from the ScanNet dataset.

\subsection{Experimental setup}

\subsubsection{Implementation}
Our final model is described in Figure~\ref{fig:system}. We sample up to 32 parts per object and 250 points per part. The representation of each object part is fed through four 1D convolutional layers (kernel size one with filters 16, 16, 32 and 256) and max-pooling is applied over the whole set of points from the object part. The feature dimension is reduced with two convolutional layer (kernel size one) of size 256 and 128. The output is then passed to four graph convolutional layers. Each of these layers has eight attention heads with respectively 16, 16, 32 and 32 filters. The output of each attention head is concatenated at each layer before being fed to the next. Finally, the resulting features are max-pooled over all object parts and the result is passed to the classification layers that consists of two fully connected layers of size 128 and 256. When predicting over single nodes, the same classification layers are applied directly on each object part representation.

\subsubsection{Datasets}
Evaluation is performed on two datasets: The ModelNet dataset~\cite{MODELNET}  and the ScanNet dataset~\cite{dai2017scannet} (1513 reconstructed rooms). We use both ModelNet40 (12311 CAD models split between 40 classes) and the ModelNet10 subset (4899 models in 10 classes). The second version of the annotation for ScanNet is used with the train/test/val split defined in the first version. Objects are extracted according to the annotation in the dataset. Afterwards the object classes are mapped to the ModelNet classes. 

\subsection{Evaluation on artificial data}
Table \ref{table:expMN40} compares the performance of our method to state-of-the-art methods on the ModelNet~\cite{MODELNET} dataset. It should be noted that the models of the dataset are non-manifold meshes. To apply our method, we project views of those objects and reconstruct them using a TSDF-based reconstruction. As a result, the object models differ slightly. For reference, we provide the accuracy of the PointNet architecture \cite{qi_pointnet:_2017} on both versions and observe a drop in accuracy of one percent for the object models created for our approach.

\begin{table}[t]
\caption{Classification accuracy on the ModelNet40 dataset \cite{MODELNET} (\textbf{*}~indicates that the method was evaluated on the reconstructed models)}
\label{table:expMN40}

\begin{center}
\begin{tabular}{|c| c| c|}
  \hline
  \textbf{Method} & \textbf{MN40} & \textbf{Input} \\
  \hline
  VoxNet~\cite{maturana2015voxnet} & 83.0 & Voxel Grid \\
  \hline
  KD-Networks~\cite{Klokov_2017_ICCV} & 91.8 & KD-Tree \\
  \hline
  MVCNN~\cite{su15mvcnn} & 90.1 & Views \\
  \hline
  MVCNN-New~\cite{10.1007/978-3-030-11015-4_49} &  \textbf{95.0} & Views \\
  \hline
  3DmFV-Net~\cite{ben20183dmfv} & 91.6 & Point Cloud \\
  \hline
  3DCapsules~\cite{DBLP:journals/corr/abs-1811-02191} & 92.7 & Point Cloud \\
  \hline
  PointNet~\cite{qi_pointnet:_2017} & 89.2 & Point Cloud \\
  \hline\hline
  \textbf{*}PointNet~\cite{qi_pointnet:_2017} & 88.1 & Point Cloud \\
  \hline
  \textbf{*}Ours (PCA-LRF) & 86.9 & Mesh \\
  \hline
  \textbf{*}Ours (Z-LRF) & 89.4 & Mesh \\
  \hline
\end{tabular}

\end{center}
\end{table}

This experiment shows that despite being specifically designed with real-world constraints in mind, our method still shows competitive results on artificial data. The results presented in Table \ref{table:expMN40} correspond to our method with a max pooling and trained with the average angle values as an included feature. Table \ref{table:expMN10design} shows the results of the addition of the average angle value as an extra feature. We see that performance slightly improves in all conditions without adding any computation as it is already calculated during the sampling process. Furthermore, the SingleNode type of pooling (i.e.~predicting a class for each object part and averaging the prediction over the object) gives similar results to MaxPool on artificial data. All future results include the average angle as a feature and use the Z-LRF.

\begin{table}[t]
\caption{Evaluation of design choices on ModelNet10}
\label{table:expMN10design}

\begin{center}
\begin{tabular}{|c |c| c| c|}
  \hline
  \textbf{Pooling} & \textbf{LRF} & \textbf{Avg ang.} & \textbf{Acc.} \\
  \hline
  MaxPool & PCA-LRF & \xmark & 85.6 \\
  \hline
  MaxPool & PCA-LRF & \cmark & 86.3  \\
  \hline
  MaxPool & Z-LRF & \xmark & 87.2\\
  \hline
  MaxPool & Z-LRF & \cmark & 89.2 \\
  \hline
  SingleNode & Z-LRF & \cmark & 89.6 \\
  \hline
\end{tabular}
\end{center}
\end{table}

\begin{figure}[t]
   \centering
   \includegraphics[scale=0.15]{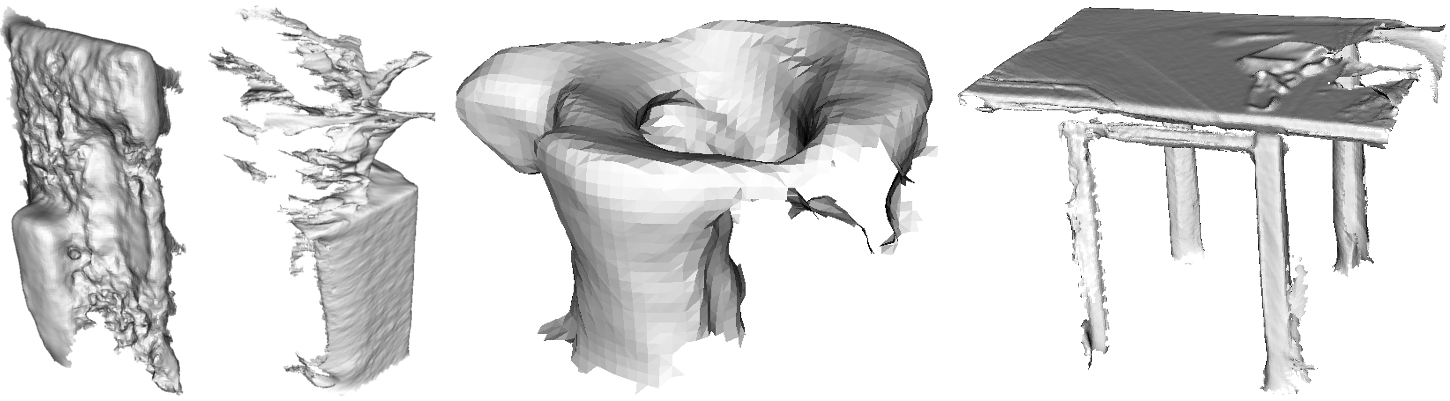}
   \caption{Illustration of the noisy objects extracted from the ScanNet dataset. Left to right: monitor (very noisy surface), potted plant (strong occlusions and many disconnected parts), cup (small object) and table (object on top merged with it).}
   \label{fig:objects}
\end{figure}

\subsection{Evaluation of the gap between real and artificial data}
This section evaluates the gap when training on artificial data (objects from ModelNet) and testing on real-world data (segmented objects from ScanNet).
This is a difficult task because of the domain shift as well as some important characteristics of the ScanNet dataset (see Figure~\ref{fig:objects}).
ScanNet was first and foremost designed for semantic segmentation, which is more concerned with large-scale structures. Additionally, since it represents natural environments, the dataset is strongly imbalanced. Chair is a highly dominating class, which is seen by the performance of the ``chair predictor'' baseline (i.e. always predicting the chair class) in Table~\ref{table:expSNToMN40}. As such, class accuracy is a more relevant metric than overall accuracy. Moreover, most structures tend to be oversmoothed by the reconstruction algorithm. This is a side-effect of the reconstruction algorithm that tries to reconcile various noisy measurements. Subtle differences also exist between ScanNet classes and ModelNet classes (e.g. a pack of bottles in ScanNet is mapped to the bottle class, whereas that class only contains single standing bottle in ModelNet). Finally, the segmentation is often inaccurate for smaller objects. This is due to the fact that scenes are first oversegmented and then clusters are annotated. Therefore, many extracted objects still include points from the surrounding elements. All these factors contribute to making this a very challenging dataset for the task.

As shown in Table ~\ref{table:expSNToMN40}, the PointNet model fails to transfer to the real-world domain, scoring only 2.2\% accuracy and 3.3\% class accuracy. In comparison, our model achieves much higher accuracy and class accuracy with scores of 34.6\% and 19.1\% respectively. The result when always predicting the chair class not only demonstrates the imbalance of the dataset but also provides a general reference for evaluating performance. The reason to focus specifically on the PointNet architecture, besides its versatility and proven reliability in various contexts \cite{Aoki_2019_CVPR, Ge_2018_CVPR}, is that it is the closest architecture to our approach. Considering that our object part representation could easily be swapped out in our pipeline, a comparison to PointNet allows for a fairer comparison of our contribution.

For our method, the best results are achieved by increasing the sampling threshold by a factor of two compared to training. Also, SingleNode pooling is used with a disconnection rate of 75\% during training. In the next section, we evaluate those design choices in more detail.

\begin{table}[t]
\caption{Evaluation when transferring from ModelNet40 to ScanNet}
\label{table:expSNToMN40}

\begin{center}
\begin{tabular}{|c| c| c| }
  \hline
           & \textbf{Acc.} & \textbf{Cls Acc.} \\
  \hline
  Chair Predictor & \textbf{36.2} & 2.5 \\
  \hline
  PointNet & 2.2 & 3.3 \\
  \hline
  Ours & 34.6 & \textbf{19.1} \\
  \hline
\end{tabular}
\end{center}
\end{table}

\subsubsection{Influence of the part size in the transfer}

\begin{figure}[t]
   \includegraphics[scale=0.5]{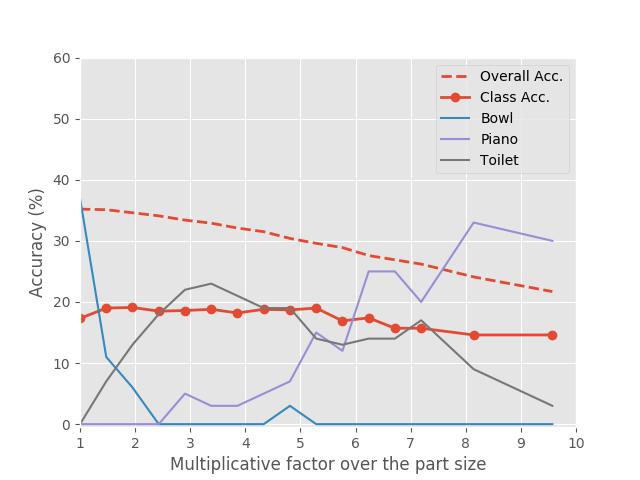}
   \caption{Accuracy in percent for various multiplicative factors applied to the sampling threshold (ModelNet40 mapping is used, SingleNode model trained with 75\% Disconnect).}
   \label{fig:size}
\end{figure}

One significant parameter in the transfer performance is the threshold set for the part sampling algorithm. ScanNet scenes tend to be oversmoothed, which affects the angle-based sampling procedure for objects that have large flat surfaces in artificial models. Also, scenes are reconstructed at a constant density, which means that smaller objects have a smaller density than bigger objects. Combined with the low level of noise, the right trade-off needs to be found for increasing the threshold used in training and testing because classes react differently. Figure \ref{fig:size} shows the impact of the sampling threshold on the accuracy when using the mapping from ModelNet40 classes to the ScanNet objects. The best threshold typically increases with the increase of the average size of the class. For small objects, such as ``bowl'', the best threshold is close to the training value. For medium-sized objects, such as ``toilet'', the best threshold is between three and fours times the original value. For very large objects, such as ``piano'', the best best threshold is up to eight times. Larger objects simply have more triangles in the mesh. The threshold is reached much faster due to the noise, therefore, they require a larger threshold in order to reproduce parts similar to those observed during training. 

\subsubsection{Influence of the connectivity and pooling}
Table \ref{table:expTransferDesign} shows the impact of different pooling strategies on the transfer performance. It also shows the impact of randomly disconnecting some object parts in the object part graph during training in order to better approximate occluded objects. The experiments are performed using the mapping from ModelNet10 to ScanNet. The significant increase in accuracy is due to the fact that the ModelNet10 mapping contains most of the well-reconstructed objects because classes of ModelNet10 corresponds to bigger structures. Clearly, performance improves with disconnection considered. Applying 0\% has low accuracy because it does not model the occlusion. At 100\% the performance is also low because this over-estimates the level of occlusion. The best compromise is achieved with 75\% for the ScanNet dataset. The disconnect experiments are not applied to the MaxPool setup, as it would still take into account each part when max-pooling, thus failing to simulate occlusions.

\begin{table}[t]
\caption{Evaluation of design choices for transferring to ScanNet (using the ModelNet10 mapping)}
\label{table:expTransferDesign}

\begin{center}
\begin{tabular}{|c| c| c| c| }
  \hline
  \textbf{Pooling} & \textbf{Disconnect (\%)} & \textbf{Acc.} & \textbf{Cls Acc.} \\
  \hline
  MaxPool & - & 44.1 & 42.35 \\
  \hline
  SingleNode & 0 & 50.1 & 38.3 \\
  \hline
  SingleNode & 50 & 62.2 & 39.0 \\
  \hline
  SingleNode  & 75 & \textbf{62.9} & \textbf{43.2} \\
  \hline
  SingleNode  & 100 & 54.5 & 33.2 \\
  \hline
  \hline
  Chair predictor & - & 56.0 & 10.0 \\
  \hline
\end{tabular}
\end{center}
\end{table}

\subsection{Evaluation on real data}
The large size of ScanNet makes it feasible to train methods on real-world data instead of artificial data. This is helpful because it can be used to establish an upper bound for the transfer to this dataset. The results in Table~\ref{table:expSNeval} are significantly higher than when transferring, which implies the existence of occlusion consistency for a given class. We, however, conjecture that this only holds for larger structures but not for smaller objects, such as household items. The results additionally show that an advantageous side-effect of our design is the ability to better learn from noisy data. Our method achieves an accuracy of 85.2\% and class accuracy of 62.8\%. The best results are achieved using the Z-LRF frame of reference. Prediction is performed for each node and averaged, and training is performed with a 75\% random chance of disconnecting two neighbors. In comparison, the PointNet architecture achieves much lower scores of 73.6\% accuracy and 52.3\% class accuracy, which is approximately 10\% less than our method. We also performed experiments with PointNet trained on objects rescaled to the unit sphere to prevent the method from taking advantage of scale information. The significant decrease in this case suggests that PointNet finds it more difficult to establish consistent shapes. This observation supports our argument that even though objects have inconsistent shapes, they always have consistent parts.


\begin{table}[t]
\caption{Evaluation when training and testing on ScanNet (using~the~ModelNet40 mapping)}
\label{table:expSNeval}

\begin{center}
\begin{tabular}{|c| c| c| }
  \hline
           & \textbf{Acc.} & \textbf{Cls Acc.} \\
  \hline
  Chair Predictor & 36.2 & 2.5 \\
  \hline
  PointNet & 73.6 & 52.3 \\
  \hline
  PointNet (rescaled obj.) & 53.4 & 17.2 \\
  \hline
  Ours (Z-LRF)  & \textbf{85.2} & \textbf{62.8} \\
  \hline
\end{tabular}
\end{center}
\end{table}

\section{Conclusion}\label{sec:conclusion}
This paper addressed the important robotics task of object classification from 3D data. We present an approach that transfers to real reconstructed objects when trained on clean CAD models only. Results show that our approach significantly outperforms state-of-the-art methods while maintaining competitive performance when training and testing on CAD models. The performance increase in the transfer is achieved through sampling object parts that are reproducible under rotation, occlusion and scale in combination with a graph-based deep learning architecture. Learning with a rotation-invariant and scale-invariant representation of parts enables objects to be recognized with significant portions of missing data.

The next step for this work is to create a large-scale dataset focused on objects rather than scenes to support further research in the task of transferring from widely available CAD models to noisy real-world data. This will be particularly important for the task of retrieving CAD models from candidate segments in noisy data. This will be highly relevant for service robotics systems that have to prepare for a variety of contexts by building a general world knowledge when deployed in user homes.






\bibliographystyle{IEEEtranS}
\bibliography{IEEEexample}





\end{document}